\pdfoutput=1
\documentclass[11pt]{article}
\usepackage{emnlp2022}
\usepackage{times}
\usepackage{latexsym}

\usepackage[T1]{fontenc}
\usepackage[utf8]{inputenc}

\usepackage{amsmath}
\usepackage{amsfonts}
\usepackage{amssymb}
\usepackage{mathbbol}
\usepackage[ruled,vlined]{algorithm2e}
\usepackage{microtype}
\usepackage{booktabs}

\usepackage{stfloats}
\usepackage{graphicx}
\usepackage{xcolor,colortbl}
\usepackage{pifont}
\usepackage{multirow}
\usepackage{soul}

\usepackage{amsmath,amsthm,amsfonts,amssymb}
\usepackage{bm}
\usepackage[ruled]{algorithm2e}
\usepackage{graphicx}
\usepackage{sidecap}
\usepackage{mathrsfs}
\usepackage{multirow, multicol}
\usepackage{subcaption}
\usepackage{caption}

\usepackage{wrapfig}
\usepackage{booktabs}

\usepackage{natbib}

\newcommand{\MI}{\text{I}}

\def\1{\bm{1}}

\def\vh{{\bm{h}}}

\def\vp{{\bm{p}}}

\def\vx{{\bm{x}}}
\def\vy{{\bm{y}}}
\def\vz{{\bm{z}}}

\DeclareMathAlphabet{\mathsfit}{\encodingdefault}{\sfdefault}{m}{sl}
\SetMathAlphabet{\mathsfit}{bold}{\encodingdefault}{\sfdefault}{bx}{n}

\DeclareMathOperator*{\argmin}{arg\,min}

\definecolor{RBRed}{rgb}{0.98,0.88,0.85}
\definecolor{RBBlue}{rgb}{0.81,0.80,0.94}
\definecolor{RBGreen}{rgb}{0.85,0.91,0.84}

\definecolor{RBRed}{rgb}{0.98,0.88,0.85}
\definecolor{RBBlue}{rgb}{0.81,0.80,0.94}
\definecolor{RBGreen}{rgb}{0.85,0.91,0.84}
\definecolor{DPGreen}{rgb}{0,0.45,0.24}

\title{TPDM: Selectively Removing Positional Information for Zero-shot Translation via Token-Level Position Disentangle Module}
\author{Xingran Chen \\
  University of Michigan\\
  \texttt{chenxran@umich.edu} \\\And
  Ge Zhang \\
  University of Waterloo\\
  \texttt{gezhang@umich.edu} \\\And
  Jie Fu\thanks{$\quad$Corresponding Author} \\
  BAAI \\
  \texttt{fujie@baai.ac.cn} \\}

\begin{document}
\maketitle
\begin{abstract}

Due to Multilingual Neural Machine Translation's (MNMT) capability of zero-shot translation, many works have been carried out to fully exploit the potential of MNMT in zero-shot translation. 
It is often hypothesized that positional information may hinder the MNMT from outputting a robust encoded representation for decoding. However, previous approaches treat all the positional information equally and thus are unable to selectively remove certain positional information. In sharp contrast, this paper investigates how to learn to selectively preserve useful positional information. 

We describe the specific mechanism of positional information influencing MNMT from the perspective of linguistics at the token level. We design a token-level position disentangle module (TPDM) framework to disentangle positional information at the token level based on the explanation. Our experiments demonstrate that our framework improves zero-shot translation by a large margin while reducing the performance loss in the supervised direction compared to previous works.

\end{abstract}

\section{Introduction}

% MNMT system has been proven to be trained on massive language pairs and achieves satisfying performance on zero-shot translation directions
MNMT systems have been proven to be capable of training on massive language pairs with a single model, and achieving satisfying performance, especially on zero-shot directions \cite{johnson-etal-2017-googles, aharoni-etal-2019-massively}. 
Recently, more efforts have been put to further improve the MNMT's performance by forcing the generation of language-agnostic representation via model modification \cite{raganato-etal-2021-empirical, gu-etal-2019-improved, liu-etal-2021-improving-zero, blackwood-etal-2018-multilingual, sachan-neubig-2018-parameter}, data augmentation \cite{wang-etal-2021-rethinking-zero, gu-etal-2019-improved, zhang-etal-2020-improving}, and regularization \cite{arivazhagan2019missing, al-shedivat-parikh-2019-consistency}. Besides, 
prior works have explored the flaws of existing MNMT models from the perspective of latent variables \cite{wang-etal-2021-rethinking-zero, gu-etal-2019-improved, liu-etal-2021-improving-zero}.

Among these works, \citet{liu-etal-2021-improving-zero} report a counter-intuitive observation that positional information may hinder the encoder from outputting language-agnostic representation due to the different word order of different source languages. To tackle the problem, they remove a residual connection within one layer of the transformer encoder to decrease the contained positional information in representation, thereby improving model performance on zero-shot translation. However, they consider positional information integratedly, leading to the following questions:
\begin{itemize}
    \item \textbf{Q1:} Does MNMT benefit from any part of the positional information?
    \item \textbf{Q2:} If the answer to \textbf{Q1} is positive, how much and which part of positional information do we need to improve the zero-shot MNMT performance?
    % \item We make an exhaustive study on the specific mechanism of positional information influencing MNMT performance.
    % \item We design an innovative token-level disentangle position module which improves the zero-shot MNMT performance by a large margin while losing less performance on supervised directions than in previous work.
    % \item We conduct detailed ablation experiments to demonstrate the improved robustness of our proposed framework, and justify that our model can selectively reserve positional information for tokens. 
\end{itemize}

% \textbf{Does MNMT benefit from part of the positional information?} To the extent that the answer is positive, \textbf{how much and which part of positional information do we need to improve the zero-shot MNMT performance?}

% The paper explores the specific mechanism of positional information influencing MNMT and proposes an interpretable solution accordingly.
This paper investigates how the positional information influences MNMT performance at the token level. We observe that the vanilla Transformer used for MNMT also learns to disentangle positional information during training, 
%while Transformer without position embedding performs poorly both on supervised and zero-shot directions and learns positional information during training. 
while the Transformer without positional embedding learns to catch positional information during training.
% MNMT benefits from positional information of sentence-ordering related tokens and some key tokens with strict order like number and punctuation. Some languages are relatively word-order free, so that need less positional information. 
We conduct further token-level experiments and propose explanation about the observations accordingly: (1) The word order of some languages is relatively free, meaning that not all positional information is necessary; (2) meanwhile, positional information is needed for sentence-level ordering, as well as for words requiring strict order to disambiguate the semantic meaning (\textit{e.g.}, number).
Based on this findings, we conjecture that the token-level positional information disentanglement, where the model can selectively remove positional information, will be better for boosting the quality zero-shot translation. To address this, we treat each token's semantic information and positional information as two separate latent variables and implement disentangled representation learning between them by minimizing their mutual information. Experiments confirm the effectiveness of the proposed approach on improving the zero-shot MNMT performance by a large margin while achieving comparative results on supervised tasks \cite{liu-etal-2021-improving-zero}. % Additionally, we find that our method can selectively reserve essential positional information. 
% From the further discussion, we found that this is because our proposed approach can selectively reserve essential positional information.
Our contributions can be summarized as follows:

\begin{itemize}
    \item We make an exhaustive study on the specific mechanism of positional information influencing MNMT performance.
    \item We design an innovative token-level disentangle position module which improves the zero-shot MNMT performance by a large margin while losing less performance on supervised directions than in previous work.
    \item We conduct detailed ablation experiments to demonstrate the improved robustness of our proposed framework, and justify that our model can selectively remove positional information for tokens. 
\end{itemize}
\section{Positional Information Impact Analysis}
\label{sec:analysis}

\begin{figure}[!tbp]
\centering
% \begin{minipage}[t]{1.0\linewidth}
% \setlength{\belowcaptionskip}{-0.1cm}  
\includegraphics[width=0.5\textwidth]{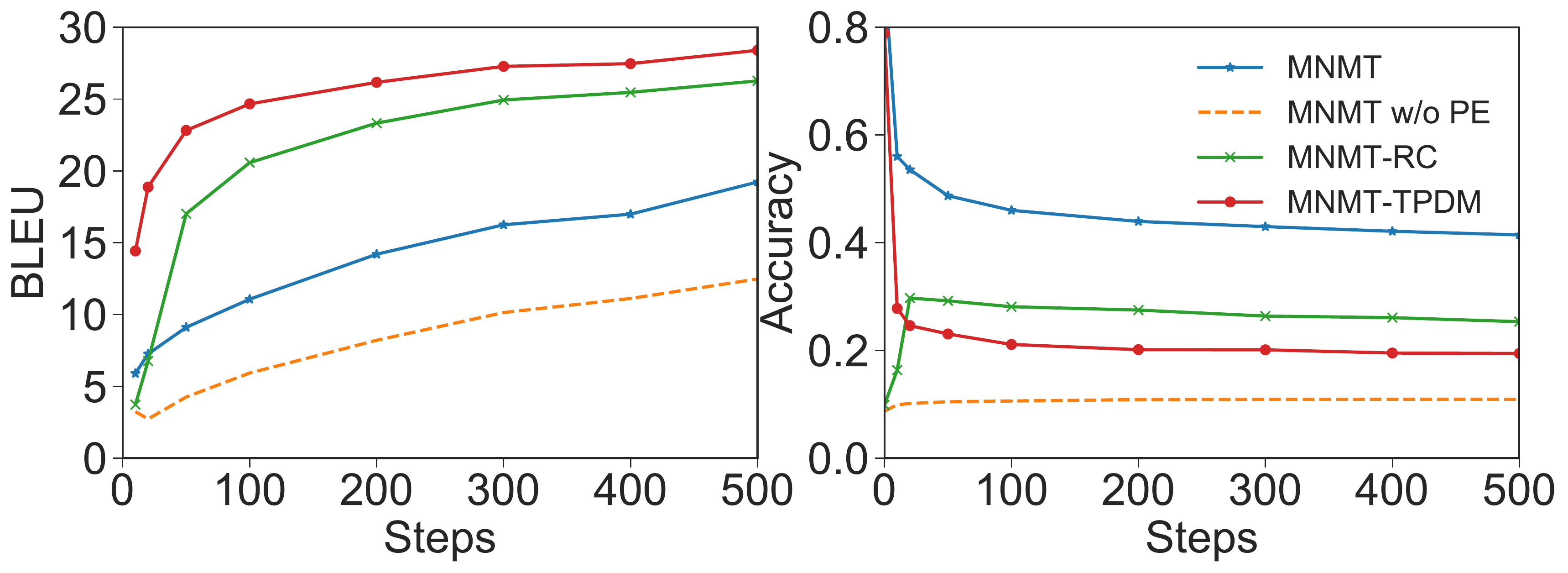}
% \label{fig:layer.1}
% \end{minipage}%}
% \hspace{0.5}
\caption{The performance curve of zero-shot translation (left) and positional information probing (right) of different MNMT models.}
\label{fig:bleu_acc}
\end{figure}

% \quad
\begin{figure}[!tbp]
% \begin{minipage}[t]{0.9\linewidth}
\includegraphics[width=0.45\textwidth]{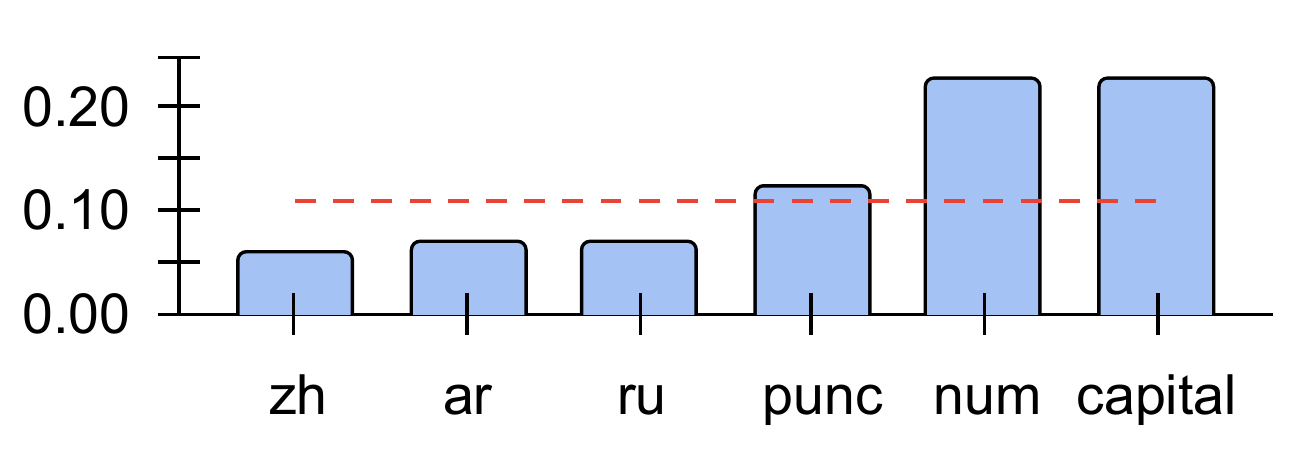}
% \label{fig:layer.2}
% \end{minipage}
% \captionsetup{font={footnotesize}} 
% \caption{\textcolor{red}{The performance of both probing task andComparison of Transformer encoder layer. The left side is the encoder layer with position disentangle module, and the right side is the original Transformer encoder layer. This figure shows that the position disentangling module is inserted right after the first residual connection within the encoder block.}}
\caption{The accuracy of MNMT without PE on the positional information probing task. The results are grouped by the different classes of tokens. The red dotted line indicates the overall accuracy of MNMT without PE on the probing task. Note that \textit{ar, zh, ru} indicate tokens of Arabic, Chinese, Russian, respectively; \textit{punc} indicates punctuation tokens, \textit{num} indicates number tokens, and \textit{capital} indicates tokens with first letter capitalized.}
\label{fig:bar}
\end{figure}

In this section, we probe the impact of positional information on translation following \cite{liu-etal-2021-improving-zero}.
Specifically, we investigate the variation of positional information within the model during training. The probing task aims to detect whether the representation of each token contains information about its corresponding position within the sentence. We choose three models for probing: 1) the vanilla Transformer (denoted as \textbf{MNMT}); 2) the residual connection cut mechanism proposed by \cite{liu-etal-2021-improving-zero} (denoted as \textbf{MNMT-RC}); 3) the vanilla Transformer without positional embedding (denoted as \textbf{MNMT without PE})\footnote{We view the baseline MNMT model as the upper bound of the positional information added to the model, and the MNMT without PE model as the lower bound of the positional information added to the model}. Experimental details are given in \S~\ref{sec:exp_detail}. During probing, we freeze all parameters of the MNMT model, and train an MLP to classify the position of each token correctly. Note that we use the zero-shot direction data to train and evaluate the classifier. The results are given in Fig.~\ref{fig:bleu_acc} and Fig.~\ref{fig:bar}.

Surprisingly, as shown in Fig.~\ref{fig:bleu_acc}, we observe that although full positional information is provided for MNMT, the accuracy of the probing task is still decreasing during training, meaning that the vanilla Transformer itself is also trying to disentangle positional information during training. Even though no positional information is provided for MNMT without PE, the accuracy of the probing task increases during training. 
These observations imply that the model keeps trying to learn positional information for achieving better performance. Additionally, MNMT without PE performs poorly both on supervised and zero-shot directions. Similarly, MNMT-RC, which achieves the best zero-shot results, tries to capture positional information in the initial stage of training. These observations indicate that MNMT benefits from optionally introducing position information. 

% Although these two phenomenons seem contradictory to each other, we consider they may be explained from the perspective of linguistics: (cite) argues that the word order of many languages ​​(e.g. German, Russian, Arabic, Polish) is relatively free when compared to other languages (e.g. English), namely that the token-level permutation within a sentence may not largely influence the semantics of the sentence; while positional information is still needed for locating key information, as well as sentence-level ordering. The results of probing also show the evidence that can justify the above opinion. In our analysis, we further calculate the probing accuracy for each token within the vocabulary, and then classify all these tokens into various classes to conduct a comparison. As shown in Figure 1b, we find that fewer ar ru tokens preserve positional information; while more punctuation, and initial capitalized tokens preserve positional information. These phenomena all show that disentangling positional information is token-specific, and the positional information that different tokens should contain should be different.

Therefore, to further understand the specific mechanism of positional information influencing MNMT performance, we calculate the probing accuracy for each token within the vocabulary and classify these tokens into various classes based on their semantics. We analyze the probing results of MNMT without PE to investigate which positional information is automatically learned to improve MNMT performance. As shown in Fig.~\ref{fig:bar}, we observe that the lower bound model 1) learns more positional information on tokens belonging to punctuation, number, and words capitalized in the first letter, but 2) preserve less positional information on Arabic, Chinese and Russian tokens. 
% We believe that MNMT benefits from positional information of key tokens and sentence-order related tokens given the first observation. The second observation matches some linguistics' research conclusion \cite{comrie1987world, rodionova2001word, ryding2005reference}, claiming that word orders of some languages, such as German, Arabic, Russian, and Polish, are relatively free.
These results indicate that disentangling positional information is achieved at the token level, and they could be well explained from the perspective of linguistics. Some linguistics research \cite{comrie1987world, alasmari2016comparative, hoffman-1996-translating} claims that the word order of some languages is relatively free compared to English. We hypothesize that MNMT should preserve positional information for tokens indicating sentence-level ordering (\textit{e.g.}, capital, punc.), and strict order phrases (\textit{e.g.}, number) to achieve better performance based on these linguistic research.

To fully utilize the token-level positional information for disentanglement, we propose a token-level position disentangle framework to help MNMT learn to disentangle positional information at the token level.

\section{Methodology}

\begin{figure*}[t]
\centering
\includegraphics[width=0.85\textwidth]{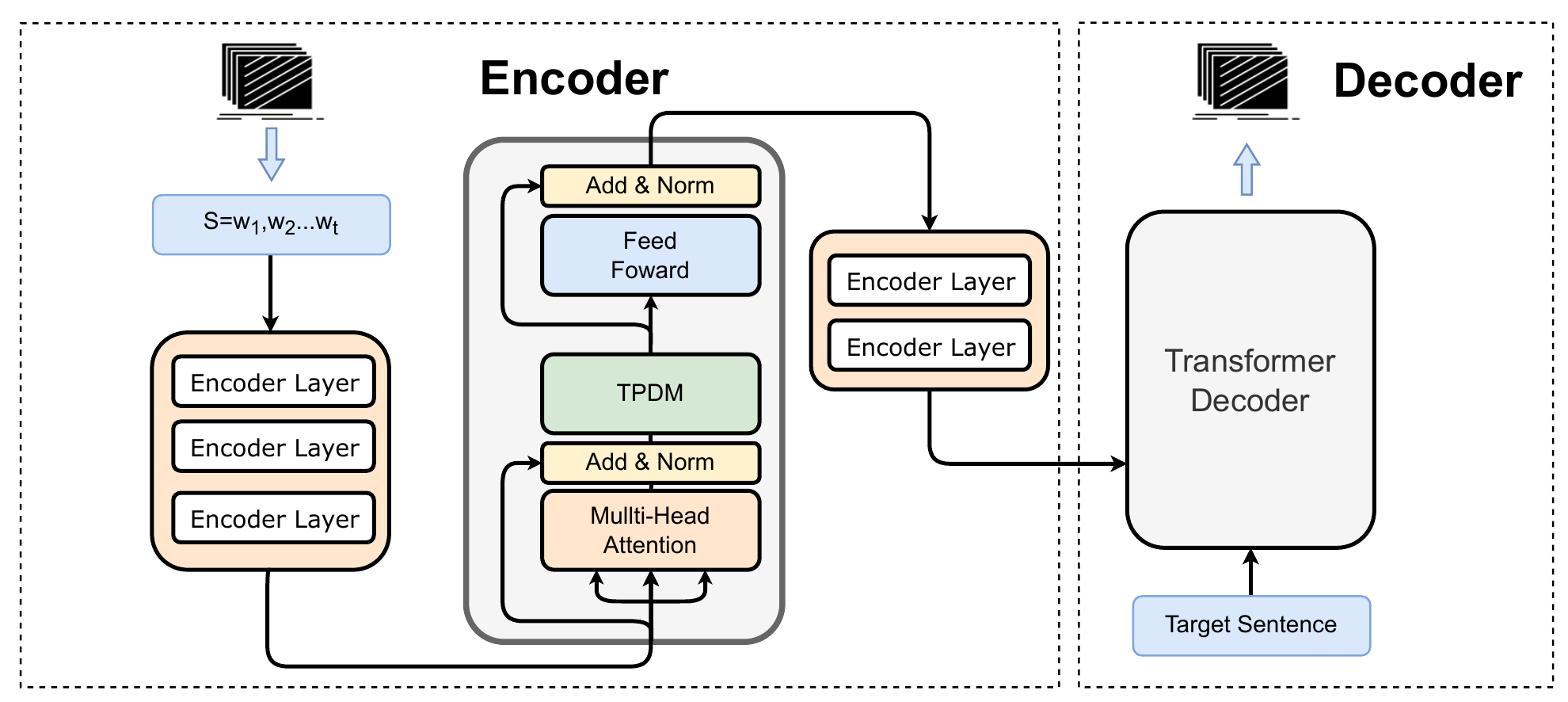}
% \vspace{-0.3in}
\caption{Illustration of our proposed framework MNMT-TPDM. We add the TPDM in an encoder layer within the Transformer to disentangle positional information at the token level. In our experiments, we select $4^{\rm th}$ for base model and $2^{\rm nd}$ for small model to add TPDM.}

% Illustration of the Transformer with position disentangle module inserted in the middle layer of $m$. In the encoder layer $m$, the $H$ and $P$ will be generated to calculate the $\mathcal{L}_d$, $\mathcal{L}_p$, and $\mathcal{L}_g$. Then $H$, instead of $H_0$ in the original Transformer, will be fed into the next module within the Transformer. }}
\label{fig:demo}
\end{figure*}

In this section, we introduce the Token-Level Position Disentangle Module (TPDM) for MNMT to address the problem we discussed in \S~\ref{sec:analysis}. We set the Transformer as our backbone, and we propose a token-level position disentangle module to add to the model. The illustration of our model is shown in Fig.~\ref{fig:demo}. In \S~\ref{sec:posDisentangleModule}, we introduce our proposed TPDM and two training objectives for the model and the adversarial generator. 
\S~\ref{sec:constraintPosHead} describes a training regularization to make sure that the generated representation of the positional head represents the positional information.

\subsection{Preliminary}
\label{sec:method:preliminary}

\subsubsection{Transformer}
% In this paper, we utilize the Transformer \cite{vaswani2017attention} as our base model to conduct experiments. Transformer consists of an encoder, which is used to encode source text into contextualized semantic representation, and a decoder, which decodes the target sentence from the representations fed from the encoder. Furthermore, both encoder and decoder are a stack containing $n$ of encoder/decoder block. To better illustrate our approach, in this subsection, we will discuss the encoder block in detail. The structure of each encoder layer is shown in Figure\ref{fig:layer}. For the $i^{th}$ encoder block, given an input hidden states $H_{i-1}$, it will first be fed into a multi-head attention module followed by a residual connection; then, it will be further fed into a feed-forward layer followed by another residual connection. \cite{liu-etal-2021-improving-zero} introduced an operation to remove the first residual connection in a middle layer $m$ of the Transformer encoder to disentangle the semantic embedding from positional information. In this paper, we propose inserting a position instead of disentangling module to allow the model to learn to disentangle positional information.

In this paper, we use the Transformer \cite{vaswani2017attention} as our baseline. The Transformer consists of an encoder and a decoder, which are stacks containing $n$ of encoder/decoder block. In this subsection, we will discuss the encoder block in detail. Given an input hidden states $H_{i-1}$, the $i^{\rm th}$ encoder block first feed its input into a multi-head attention module followed by a residual connection; then, it will be further fed into a feed-forward layer followed by another residual connection. \cite{liu-etal-2021-improving-zero} introduced an operation to remove the first residual connection in a middle layer $m$ of the Transformer encoder to disentangle the semantic embedding from positional information. We propose TPDM to allow the model to learn to disentangle positional information at the token level.

\subsubsection{Disentangle Representation Learning via Minimizing Mutual Information}
\label{sec:disentangle preliminary}
% In disentangling representation learning, given two latent embeddings, $\vy$ and $\vz$ drawn from the distribution $q_\theta(\vy, \vz|\vx)$ can be reduced to minimizing the mutual information $\MI(\vy;\vz)$ between them. However, finding a trainable estimator for $\MI(\vy;\vz)$ remains hard. Recently, \cite{cheng-etal-2020-improving} presents an adversarial training-based approach to minimize mutual information and then minimize this upper bound instead during training. The sample-based mutual information upper bound is:

In disentangling representation learning, given two latent embeddings $\vy$ and $\vz$ drawn from the variational distribution $q_\theta(\vy, \vz|\vx)$, disentangling $\vy$ and $\vz$ can be achieved by minimizing the mutual information $\MI(\vy;\vz)$ between them~\cite{cheng-etal-2020-improving}. 
However, obtaining a differentiable estimation of mutual information for back-propagation could be intractable \cite{cheng2020club}, and inaccurate when the dimension increases \cite{belghazi2018mutual}. Recently, \cite{cheng-etal-2020-improving} proposes an adversarial training-based approach to minimize the upper bound of mutual information instead to decrease the computational difficulty. The proposed sample-based mutual information upper bound is:

\begin{equation}
\small
\label{eq:mi}
\hat{\MI}(\vy;\vz) = \frac{1}{M}\sum_{j=1}^M \left \{\log p_\sigma(\vz_j|\vy_j) - \frac{1}{M} \sum_{k=1}^M \log p_\sigma(\vz_j|\vy_k)\right \}, %E[\frac{1}{M}\sum_{j=1}^M log p_\sigma(h_2|h_1)- E_c[log p_\sigma(h_j|p_k)]]
\end{equation}
% $p_\sigma$ is a generator that learns the mapping from $\vy$ to $\vz$ through maximizing $\log p(\vy|\vz)$. In this case, the problem of minimizing mutual information can be reduced to minimizing the sample-based upper bound $\hat{\MI}(\vy;\vz)$, and therefore the disentangle representation learning objective for $q_\theta$ is:
where $p_\sigma$ is a generator that learns the mapping from $\vy$ to $\vz$ through maximizing $\log p(\vy|\vz)$. Given the upper bound, the disentangle representation learning objective for $q_\theta$ becomes:

\begin{equation}
\theta^* = \argmin_\theta \hat{\MI}_{\theta}(\vy;\vz).
\end{equation}

Note that the $p_\sigma$ and $q_\theta$ will be updated alternatively during training. In this paper, we utilize the method to disentangle the positional embedding. First, we treat the encoder of the Transformer as $q_\theta$, then we generate two latent embeddings $\vh_i$, $\vp_i$ representing semantic and positional information from $q_\theta$, to conduct disentangle representation learning.

\subsection{Multilingual Neural Machine Translation System}

\begin{figure}[!tbp]
% \begin{minipage}[t]{0.45\linewidth}
\centering
% \vspace{-0.45in}
\includegraphics[width=2.2in]{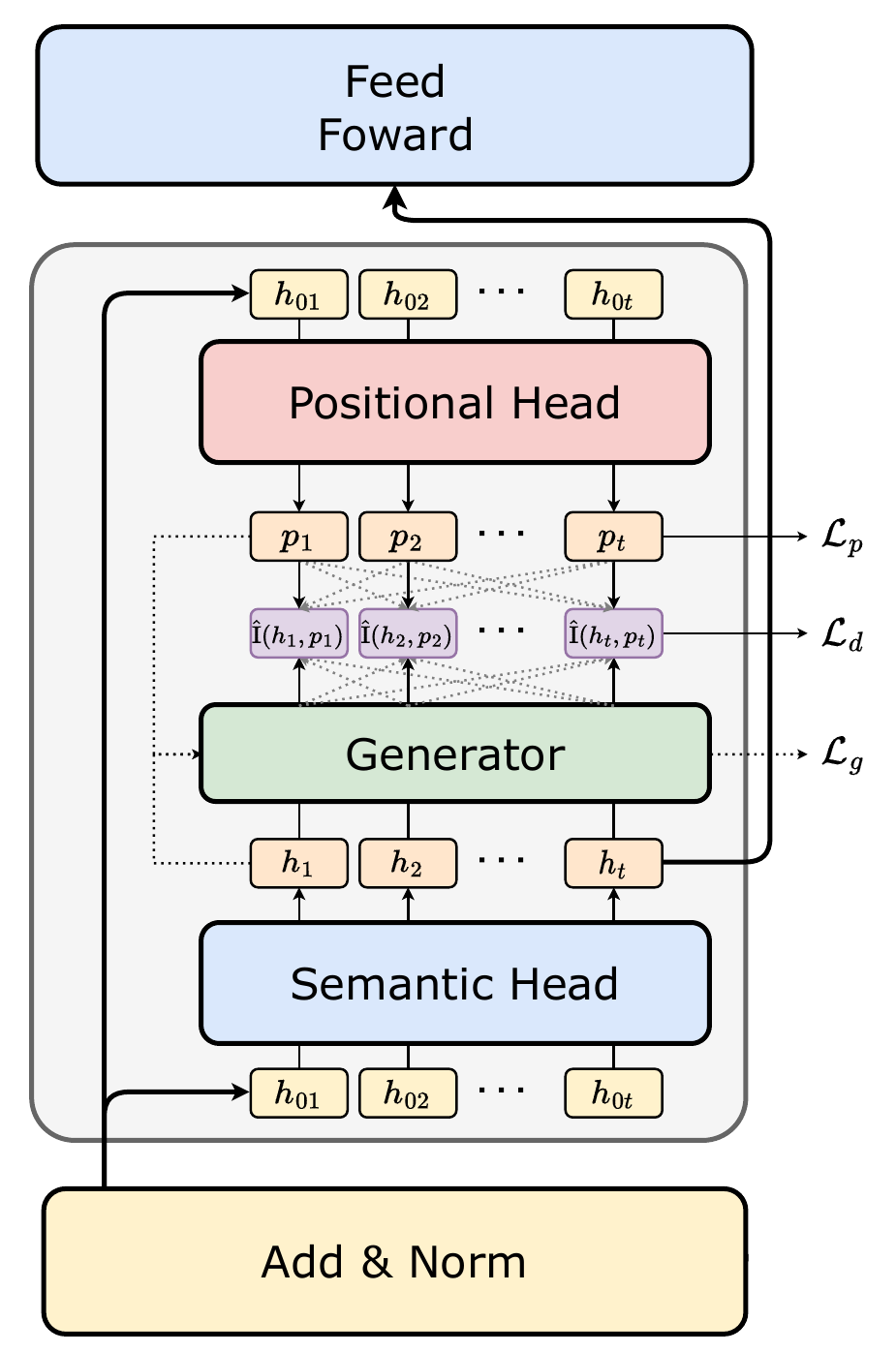}
% \setlength{\belowcaptionskip}{-0.1cm}  
% \includegraphics[height=5.0cm,width=3.5cm]{figs/Transformer_dp.pdf}
% \end{minipage}%
% \begin{minipage}[t]{0.45\linewidth}
% \centering
% \includegraphics[height=3.9cm,width=3.5cm]{figs/Transformer.pdf}
% \end{minipage}
% \captionsetup{font={footnotesize}} 

\caption{Detailed illustration of the structure of TPDM in Fig.~\ref{fig:demo}. The input is fed into two Siamese MLP heads to generate semantic latent embedding $\vh_i$ and positional latent embedding $\vp_i$ for each token, respectively. Then token-level mutual information $\hat{\MI}(\vh_i;\vp_i)$ is calculated for disentangling the model. The latent semantic embedding $\vh_i$ is then viewed as the output of the module to be fed into the feed-forward module.}

\label{fig:layer}
\end{figure}

Given a multilingual parallel dataset $\{(\vx_k, \vy_k, s_k, t_k)\}_{k=1}^D$, where $\vx_k$ represents the source text, $\vy_k$ represents the target text, and $s_k$, $t_k$ represents the language ID of $\vx_k$ and $\vy_k$, respectively, training a MNMT model with parameters $\theta$ involves optimizing the following training objective, where $x^*_k$ is the original sentence started with target language ID token <$t_k$> referring to the initial setting of \cite{johnson-etal-2017-googles}:

\begin{equation}
    \mathcal{L}(\theta) = -\sum_{k=1}^D \log p(\vy_k|x^*_k; \theta).
\end{equation}
%where $s_i$ is represented by a language ID token within the model in the initial setting of \cite{johnson-etal-2017-googles}.
 % An important and valuable discovery on MNMT is that it is capable of conducting zero-shot direction translation, which is unseen during training and it is therefore valuable for low resource language. This paper focuses on improving the zero-shot translation performance for the MNMT system.

\subsection{Token-Level Position Disentangle Module (TPDM)}
\label{sec:posDisentangleModule}

We propose a position disentangle module to disentangle positional information at the token level. The detailed illustration of our proposed module is shown in Fig.~\ref{fig:layer}. 
% We will explain the specific mechanism of the proposed module as follow. 

% Here we discuss how the position disentangle module works within the Transformer encoder. 
% \textcolor{red}{Here we discuss how the proposed module disentangle positional information for each token within the Transformer encoder.} Given a sentence $\vx=\{x_1,...,x_t\}$, \textcolor{red}{where $x_i$ represents each token within the sentence,} we first view the whole encoder as $q_\theta$, and then draw both semantic latent embedding $\mH=\{\vh_1,..,\vh_t\}$ and position latent embedding $\mP=\{\vp_1,...,\vp_t\}$ from it. Inspired by \cite{li2021disentangled}, we propose heads $\mathcal{M}_h$ and $\mathcal{M}_p$, which are \textbf{not weight shared}, to generate $H$ and $P$ respectively from the encoder. To be specific, given a hidden states $\mH_0=\{\vh_{01},...,\vh_{0t}\}$, we have:

% \begin{equation}
%     \mH = \mathcal{M}_h(\mH_0),
% \end{equation}

% \begin{equation}
%     \mP = \mathcal{M}_p(\mH_0).
% \end{equation}

Given a token $x$ within a sentence $\vx=\{x_1,...,x_t\}$, we first view the whole encoder as $q_\theta$, and then draw both semantic latent embedding $\vh_i$ and positional latent embedding $\vp_i$ of token $x_i$ from it. Since the Transformer encoder can only output a single embedding, to tackle with this problem, inspired by \cite{li2021disentangled}, we propose two siamese MLP heads $\mathcal{M}_h$ and $\mathcal{M}_p$, which are \textbf{not weight shared}, to generate $\vh_i$ and $\vp_i$ respectively. To be specific, given a hidden states $\vh_{0i}$, we have:

\begin{equation}
    \vh_i = \mathcal{M}_h(\vh_{0i}),
\end{equation}
\begin{equation}
    \vp_i = \mathcal{M}_p(\vh_{0i}).
\end{equation}

% Note that as shown in Fig.~\ref{fig:demo}, we refer to \cite{liu-etal-2021-improving-zero} to set the hidden state right after the first residual connection within the middle layer of the encoder as $\vh_{0i}$. 
We aim to minimize the mutual information $\MI(\vh_i, \vp_i)$ to disentangle $\vh_i$ and $\vp_i$, and Eq.~\ref{eq:mi} is used to calculate the estimation of $\MI(\vh_i, \vp_i)$. Specifically, we define a neural network $\mathcal{G}_\sigma$ with parameters $\sigma$ to represent the generator $p_\sigma$. Then, we project $\vh_i$ to the embedding space of $\vp_i$ utilizing $\mathcal{G}_\sigma$.
% \begin{equation}
%     \hat{p}_i = \mathcal{G}(h_i)
% \end{equation}
Finally, we calculate the conditional probability $p_\sigma(\vp|\vh)$ as follow:

\begin{equation}
\label{eq:conditional_probability}
    p_\sigma(\vp_j|\vh_i) = \frac{sim(\mathcal{G}(\vh_i), \vp_j) / \tau}{\sum_{k=1}^t sim(\mathcal{G}(\vh_i), \vp_k) / \tau},
\end{equation}

where $\vp_k$, $\vp_j$ belong to tokens $x_k$, $x_j$ within the sentence, respectively. $sim(\cdot)$ represents the cosine similarity function, and $\tau$ represents the temperature hyper-parameter. 
During training, given a mini-batch of data $\{(\vx_i, \vy_i)\}_{i=1}^b$, we calculate the estimation of mutual information $\hat{\MI}(\vh_{ij};\vp_{ij})$ of each source token $x_{ij}$ respectively. Hence, the disentangle training objective for the whole model, and the training objective for the generator $\mathcal{G}_\sigma$ are:

\begin{equation}
    \label{eq:ld}
    \mathcal{L}_d(\theta) = \sum_{i=1}^b \sum_{j=1}^t \hat{\MI}(\vh_{ij};\vp_{ij}),
\end{equation}
\begin{equation}
    \label{eq:lg}
    \mathcal{L}_g(\sigma) = -\sum_{i=1}^b \sum_{j=1}^t \log p_\sigma(\vp_{ij}|\vh_{ij}).
\end{equation}

As shown in Fig.~\ref{fig:layer}, the semantic latent embedding $h_i$ of each token  will be fed into the feed-forward layer within the encoder layer.
\subsection{Regularization of Positional Embedding Head}
\label{sec:constraintPosHead}
In this subsection, we introduce a regularization for positional head $\mathcal{M}_p$ to ensure the generated latent positional embedding $p_i$ of $\mathcal{M}_p$ could correctly represent the position of the corresponding input tokens. To address this, an intuitive consideration is that the positional embedding should be example-agnostic. 
Hence, we follow \cite{gao-etal-2021-simcse, chen2020simple} to maximize agreement among embeddings in the same positions while pushing those that are in different positions away. Formally, given a mini-batch of training examples $\{\vx_i, \vy_i\}_{i=1}^b$, we set the auxiliary training objective for position alignment as follows:
\begin{equation}
\label{eq:lp}
    \mathcal{L}_p(\theta) = \sum_{i=1}^b\sum_{j=1}^t - \log \frac{sim(\vp_{ij}, \bar{\vp}_j) / \tau}{\sum_{k=1}^t sim(\vp_{ij}, \bar{\vp}_k) / \tau}
\end{equation}
where $\bar{\vp}_k = \frac{1}{b}\sum_{i=1}^{b}\vp_{ik}$ represent the contrastive learning anchor of the $k^{\rm th}$ position, $\tau$ represents the temperature hyper-parameter.

\subsection{Gated Mechanism for Automatic Token-Level Disentanglement}
We also try a variant of TPDM, where we introduce a gated mechanism to implement the selective mechanism. To be specific, inspired by \cite{paranjape-etal-2020-information}, we introduce a learnable mask $\vz$ for the MNMT to selectively mask the unnecessary mutual information:

\begin{equation}
   \hat{\MI}^{'}(\vh_{i};\vp_{i}) = \vz \otimes \hat{\MI}(\vh_{i};\vp_{i}),
\end{equation}
where $\vz \in {\{0, 1\}}^t$, and $t$ is the length of input sentence. To obtain a differentiable $\vz$ for model to learn to disentangle at the token-level, we apply a differentiable and continuous binary mask $\vz^*$ proposed by \cite{paranjape-etal-2020-information} to approximate $\vz$. Therefore, the modified training objective for disentangling becomes:

\begin{equation}
    \label{eq:final_ld}
    \mathcal{L}^{'}_d(\theta) = \sum_{i=1}^b \sum_{j=1}^t \vz^* \otimes \hat{\MI}(\vh_{ij};\vp_{ij}),
\end{equation}

\subsection{Training Objectives}
Overall, the training objective of the Transformer model with TPDM is to find parameters $\theta^*$ such that:
\begin{equation}
\label{eq:overall}
    \theta^*=\argmin_\theta  \left( \mathcal{L}(\theta) + \mathcal{L}_d(\theta) + \mathcal{L}_p(\theta) \right),
\end{equation}

Note that when the gated mechanism is applied, the $\mathcal{L}_d(\theta)$ in Eq.~\ref{eq:overall} will be replaced by $\mathcal{L}^{'}_d(\theta)$.

The training objective for the adversarial generator $\mathcal{G}$ is to find parameters $\sigma^*$ such that:
\begin{equation}
    \sigma^* = \argmin_\sigma \mathcal{L}_g(\sigma).
\end{equation}

Following \cite{cheng-etal-2020-improving}, for each step in the training process, we first calculate Eq.~\ref{eq:lg} and update the parameter of $\mathcal{G}$, then we utilize $\mathcal{G}$ to re-compute the $p_\sigma(\vp_{ij}|\vh_{ij})$ to calculate Eq.~\ref{eq:final_ld}. Finally, we use Eq.~\ref{eq:overall} to update the MNMT model.

\section{Datasets and Experiment Setting}
% \textcolor{red}{We study the translation performance of our model on two datasets and measure the adaptability of our tricks with two other existing works.}
This section introduces the selected datasets and the experiment setting.

\subsection{Datasets}
\textbf{MultiUN} is a widely-used translation corpus from the United Nations \cite{eisele-chen-2010-multiun}. We select Arabic (Ar), Chinese (Zh), Russian (Ru), and English (En) for our experiment. The Ar, Ru, Zh $\leftrightarrow$ En data are used to train the model, and the Zh $\leftrightarrow$ Ar, Ar $\leftrightarrow$ Ru, Zh $\leftrightarrow$ Ru data are used for evaluating the performance of the models on zero-shot direction translation. We select these three languages (Ar, Ru, Zh), because they represent different language families, increasing the difficulty of the zero-shot MNMT task. Note that the MultiUN is also used for experiments in \S~\ref{sec:analysis}.

% Following \cite{wang-etal-2021-rethinking-zero}, we use the standard validation set of MultiUN to achieve model selection and evaluation.

\textbf{Europarl} is a well-known machine translation parallel corpus \cite{koehn2005europarl} extracted from the proceedings of the European Parliament. Here we choose German (De), French (Fr), and English (En) for performance measurement. We use De $\leftrightarrow$ En and Fr $\leftrightarrow$ En data to train the model, and Fr $\leftrightarrow$ De is used for the zero-shot direction evaluation.

\begin{table*}[ht]
\centering
\resizebox{\textwidth}{!}{%
\begin{tabular}{@{}lcccccccc|cccc@{}}
\toprule
\textbf{Task}                  & \multicolumn{8}{c}{\cellcolor{RBRed}MultiUN} &  \multicolumn{4}{c}{\cellcolor{RBBlue}Europarl} \\ \midrule & 
Ar$\rightarrow$Zh  & Zh$\rightarrow$Ar & Ar$\rightarrow$Ru    & Ru$\rightarrow$Ar   & Zh$\rightarrow$Ru    & Ru$\rightarrow$Zh      & \textbf{Zero. Avg.} & \textbf{Sup. Avg.} & De$\rightarrow$Fr  & Fr$\rightarrow$De  & \textbf{Zero. Avg.} & \textbf{Sup. Avg.}     \\ \midrule
% \multicolumn{12}{l}{\textbf{Small Model}}   \\ \midrule
PIV-M (small) & 35.0 & 17.6  & 24.0 & 19.4  & 20.9 & 33.7 & 25.1 & 42.8 & 24.7  & 20.0 & 22.4 & 30.6    \\ 
PIV-M (base)  & 42.8 & 22.9  & 30.4   & 23.3  & 28.1   & 40.8 & 31.4 & 48.3 & 28.0  & 22.3 & 25.2 & 33.5    \\ \midrule
MNMT-small      & 7.2   & 4.7   & 4.1   & 5.4 & 3.9 & 7.1 & 5.4 & \textbf{42.8} & 15.3 & 9.4 & 12.4 & \textbf{30.6}     \\
MNMT-RC-small   & 10.3  & 6.9   & 4.2   & 6.5 & 2.8 & 10.5 & 6.9   & 42.5 & 20.6 & 4.5 & 12.6 & 30.4   \\
MNMT-TPDM-small   & 26.2 & 13.2  & 18.7   & 14.3  & 16.6  & 26.0     & 19.2 & 42.3 & 21.8 & 17.9 & 19.9 & 30.5  \\
MNMT-GTPDM-small   & \textbf{28.8} & \textbf{15.0}  & \textbf{19.8}   & \textbf{16.3}  & \textbf{17.3}  & \textbf{29.4}     & \textbf{21.1} & 42.1 & \textbf{23.7} & \textbf{19.1} & \textbf{21.4} & 30.0  \\ \midrule 
% & \multicolumn{12}{l}{\textbf{Base Model}}   \\ \midrule  
% & \textbf{Supervised Avg.}  & ar-zh  & zh-ar & ar-ru    & ru-ar   & zh-ru    & ru-zh      & \textbf{Zero-shot Avg.}  & \textbf{Supervised Avg.} & de-fr  & fr-de  & \textbf{Zero-shot Avg.}      \\ \midrule
MNMT-base           & 29.3 & 15.0  & 20.9   & 16.2  & 18.6   & 29.4 & 21.6 & \textbf{48.3} & 27.4  & 21.2 & 24.3 & \textbf{33.5}    \\
MNMT-RC-base        & 41.0 & 21.0  & 29.1  & 21.7   & 26.2   & 40.0 & 29.8 & 47.4 & 29.3 & 23.2 & 26.3 & 32.9  \\
MNMT-TPDM-base        & \textbf{44.0} & \textbf{22.4}  & 30.7   & 23.1  & \textbf{27.7}  & 42.7     & \textbf{31.8} & 47.8 & 29.2 & 24.0 & 26.6 & 33.0 \\
MNMT-GTPDM-base   & 43.7 & 22.3  & \textbf{30.8}   & \textbf{23.4}  & 27.4  & \textbf{43.1}     & \textbf{31.8} & 47.4 & \textbf{29.9} & \textbf{24.3} & \textbf{27.1} & 32.7  \\ \bottomrule
\end{tabular}%
}
\caption{Experimental results on MultiUN dataset and Europarl dataset. The symbol $a \rightarrow b$ indicates the performance of models on translating language $a$ to language $b$. \textbf{Zero. Avg.} represents the average results on all zero-shot direction data; \textbf{Sup. Avg.} represents the average results on all supervised-direction data. PIV-M indicates the performance of pivot-based translation using MNMT.}
\label{tab:main_results}
\end{table*}

\subsection{Baselines}
We denote our approach as \textbf{MNMT-TPDM}, and the gated variant as \textbf{MNMT-GTPDM}. In addition, we compare with the following baselines: First, we follow \cite{johnson-etal-2017-googles} to set a baseline trained on a vanilla Transformer denoted as \textbf{MNMT}. \textbf{PIV-M}~\cite{gu-etal-2019-improved} represents pivot-based translation using MNMT. In addition, we compare our approach with removing residual connection operation proposed by \cite{liu-etal-2021-improving-zero}, which we denoted as \textbf{MNMT-RC}. 

% Furthermore, we also conduct our experiments on denoising setting \textbf{(DN)} \cite{wang-etal-2021-rethinking-zero} in Sec \ref{sec:discussion}, where a denoising autoencoder task is defined to strengthen the model's ability to detect target language ID. We re-implement all these baselines in our setting for a fair comparison. We also propose other analyses concerning our approach to conduct a comprehensive investigation.

% All baseline methods mentioned are re-implemented and followed the same experiment setting with our methods to confirm a fair comparison. Additionally, we propose the experiments of combining methods mentioned with our tricks to demonstrate the adaptablity of our model.   

\subsection{Experimental Details}
\label{sec:exp_detail}
In this subsection, we will discuss our experimental settings in detail.  
% Our experimental setting is following \cite{wang-etal-2021-rethinking-zero}.

\subsubsection{Transformer} In our experiments, we use the same setting as \cite{vaswani2017attention} to set the number of Transformer layer $n_{layer} = 6$, attention head $n_{head} = 8$, and size of dimension $d = 512$. For a comprehensive comparison, we also train a smaller scale of model with $n_{layer} = 3$, $n_{head} = 4$ and $d = 256$. We denote the former one as \textbf{base model}, and the latter one as \textbf{small model}. Note that we use \textbf{base model} to conduct experiments for analysis in \S~\ref{sec:analysis}.

\subsubsection{TPDM} Following \cite{liu-etal-2021-improving-zero}, we choose a middle layer of Transformer to insert proposed TPDM. For the base model, we select the $\rm 4^{th}$ encoder block, while for the small model, we choose the $\rm 2^{nd}$ encoder block to insert the TPDM. During our implementation, we use MLP to implement $\mathcal{M}_h$ and $\mathcal{M}_p$.
% , and we use a Transformer encoder layer to implement generator $\mathcal{G}$.
In terms of $\mathcal{G}$, we additionally set an encoder layer of Transformer to contextually learn the mapping from $\vp$ to $\vh$ for each token within a sequence.

\subsubsection{Hyper-parameter Setting} We apply the same hyperparameter settings for both the small and base models. We follow \cite{wang-etal-2021-rethinking-zero} to first tokenize all the data using the \texttt{mose} package \footnote{https://github.com/moses-smt/mosesdecoder} (for Chinese, we use the \texttt{Jieba} \footnote{https://github.com/fxsjy/jieba} toolkit), and then apply Byte Pair-Encoding (BPE) \cite{sennrich-etal-2016-neural} for 40K operations. We select Adam as our optimizer and choose learning rate 5e-4, max tokens 8000, warm-up steps 4000, and temperature $\tau = 0.05$ during training to train our model. We update our model for 300k steps on the Europarl dataset, and 500k steps for the MultiUN dataset. During the evaluation, we apply beam search = 5 to generate the target sentence and apply \texttt{SacreBLEU} \cite{post-2018-call} as our evaluation metrics to evaluate the models. We use 4 Tesla V100 GPUs to conduct our experiments.
\section{Evaluation}

\subsection{Results}

As shown in table~\ref{tab:main_results}, our proposed framework observably improves the performance on zero-shot translation. Specifically, the MNMT-TPDM and MNMT-GTPDM significantly outperform the baseline model MNMT, and gains 5.5 and 6.5 BLEU points improvement on average over the MNMT-RC. Note that the performance of MNMT-GTPDM-small is even competitive with the MNMT-base model on MultiUN. Compared to PIV-M, the MNMT-TPDM and MNMT-GTPDM model gains 0.9 and 1.2 BLEU points improvement on average under the base model setting on zero-shot directions. Empirically, we verify that our approach is capable of improving the MNMT's performance on zero-shot translation.

% \textcolor{red}{When designing ablation studies and examining the model outputs, off-target translation problems and severe grammar position bias are detected, in line with our inspiration and prior works \cite{liu-etal-2021-improving-zero} \cite{wang-etal-2021-rethinking-zero}.} 

Furthermore, our model achieves significant performance improvement on Fr $\rightarrow$ De direction in Europarl, and Ar $\rightarrow$ Zh, Ru $\rightarrow$ Zh translation directions in MultiUN. The finding shows that the MNMT-TPDM has substantial effects on different kinds of directions, even for language pairs that come from different language families. % Furthermore, we find out that although both MNMT and MNMT-RC perform poorly on several directions, especially on small model (e.g. Ar $\rightarrow$ Ru and Fr $\rightarrow$ De).
Besides, we notice that although we run for different random seeds, both MNMT and MNMT-RC perform poorly on several directions, especially for the small model.
% We run experiments with small models using different random seeds. 
% MNMT-RC, with the small models,

We find out that this is because MNMT-RC-small often translates into a wrong language (\textit{e.g.}, English), while MNMT-TPDM rarely does. This observation shows that our proposed framework is more robust, especially when the models are small.
% Additionally, we find out that the MNMT-RC frequently translates into wrong languages (e.g. English), especially on small model, in some zero-shot directions while MNMT-TPDM rarely does, indicating that our proposed MNMT-TPDM is more stable for zero-shot translation. We repeatedly ran the experiments both on small models and base models to demonstrate the robustness and reliability of the performance analysis. 

% This implies underlying practical improvements on much more unrelated and low-resource language translation directions. 

As for supervised directions translation, we find out that the degree of performance loss of our proposed MNMT-TPDM on supervised direction is further eased compared to MNMT-RC. One possible explanation can selectively preserve positional information for some key tokens when disentangling. We will discuss this in detail in \S~\ref{sec:token-level_disentangleInspection}.

In terms of the performance between TPDM and GTPDM, we find out that although the GTPDM significantly outperforms the TPDM on small model, it performs competitive when the model becomes larger. We consider that this is because the MNMT-TPDM-base model itself is capable of selectively reserving positional information. We will discuss this in detail in \S~\ref{sec:token-level_disentangleInspection}. Moreover, in supervised direction, GTPDM suffers a more severe performance loss than TPDM. We consider it is reasonable since the gated mechanism~\cite{paranjape-etal-2020-information} introduces an additional training objective for controlling the masks.
% the performance loss is also slightly eased.

% Contrary to the significant gains in the zero-shot directions, the baseline model slightly outperforms our model given a similar scale of parameters. It is understandable because the positional information we split can provide grammar hints to supervised direction translation. Meanwhile, the degree of the performance loss in the supervised direction is acceptable. Furthermore, compared to the model proposed by \cite{liu-etal-2021-improving-zero}, the performance loss is also slightly eased.

% Additionally, our model also shows favourable adaptability and robustness. Key mechanism proposed by the paper can be combined with other existing research work focusing on improving zero-shot MNMT task performance. We also successfully weakens the positional correspondence as our motivation, which will be further discussed in the following positional correspondence inspection section.

\subsection{Discussions}
\label{sec:discussion}
In this subsection, we conduct comprehensive experiments to investigate the effectiveness of our proposed framework in various settings.

\subsubsection{Ablation Studies}

We first conduct several ablation studies to investigate the importance of each component in our proposed TPDM. To be specific, we try to remove $\mathcal{L}_d$ and $\mathcal{L}_p$ from the training process, and evaluate the model's performance. Note that $\mathcal{L}_g$ does not influence any parameter within the Transformer model, therefore we do not conduct ablation studies on it. Besides, we try to utilize $i^{\rm th}$ sinusoidal positional embedding within the vanilla Transformer as the contrastive anchor $\bar{\vp}_i$ to compute $\mathcal{L}_p$. The results is shown in Table~\ref{tab:ablation_study_position}.

\begin{table}[t]
\centering
\resizebox{0.4\textwidth}{!}{%
\begin{tabular}{@{}lccc@{}}
\toprule
\textbf{Task} & \multicolumn{3}{c}{\textbf{Europarl + Small Model}}
\\ \midrule 
 & De$\rightarrow$Fr & Fr$\rightarrow$De & \textbf{Zero. Avg.}
 \\ \midrule
 MNMT-TPDM w/o $\mathcal{L}_d$ & 20.0 & 16.0 & 18.0 \\
 MNMT-TPDM w/o $\mathcal{L}_p$ & 20.6 & 15.6 & 18.1 \\
 MNMT-TPDM + sinusoidal anchor & \textbf{22.2} & \textbf{18.0} & \textbf{20.1} \\
 MNMT-TPDM & 21.8 & 17.9 & 19.9 \\
%  MNMT-GTPDM & \textbf{23.7} & \textbf{19.1} & \textbf{21.4} \\
\bottomrule
\end{tabular}%
}
\caption{Ablation studies of our proposed framework.}
\label{tab:ablation_study_position}
\end{table}

% We select anchor position representation and propose positional loss to serve the functionality of disentangling the positional information. Ablation experiments are shown in this section to verify the rationalization of the proposed vital mechanism.
As shown in Table~\ref{tab:ablation_study_position}, we can find out that removing $\mathcal{L}_d$ or $\mathcal{L}_p$ from the training process will lead to a decrease in the model's performance. The observation verifies that both $\mathcal{L}_d$ and $\mathcal{L}_p$ is crucial to gain a better performance, and it is crucial to disentangle semantic embedding from \textbf{positional information}, but not any other latent embedding. Besides, MNMT-TPDM and MNMT-TPDM + sinusoidal anchor show competitive results, meaning that our approach performs well as long as the contrastive anchors represent the corresponding position correctly.

\subsubsection{Learnable Positional Embedding Scenario}

\begin{table}[t]
% \small
\centering

\resizebox{0.45\textwidth}{!}{%
\begin{tabular}{@{}lcccc@{}}
\toprule
\textbf{Task} & \multicolumn{4}{c}{\textbf{Europarl + Small Model}}
\\ \midrule 
 & De$\rightarrow$Fr & Fr$\rightarrow$De & \textbf{Zero. Avg.} & \textbf{Sup. Avg.}
 \\ \midrule
 MNMT & 15.3 & 9.4 & 12.4 & \textbf{30.6} \\
 MNMT-TPDM  & 21.8 & 17.9 & 19.9 & 30.5 \\
 MNMT + learnable pos. & 9.1 & 7.5 & 8.3 & 30.5 \\
 MNMT-TPDM + learnable pos. & \textbf{22.7} & \textbf{18.7} & \textbf{20.7} & 30.0 \\
\bottomrule
\end{tabular}%
}
\caption{Results of learnable positional embedding scenario}
\label{tab:ablation_study_LearnedPosition}
\end{table}

We also investigate our proposed framework's effectiveness in the scenario where we follow BERT~\cite{devlin-etal-2019-bert} to replace sinusoidal positional embedding in the Transformer with a learnable positional embedding. The result in table~\ref{tab:ablation_study_LearnedPosition} show that, although the learnable positional embedding baseline reaches a competitive result on supervised-direction translation compared to the fixed counterpart, it cannot generalize well to the zero-shot direction translation task. Our proposed MNMT-TPDM reaches better performance in the zero-shot direction translation. These experimental results clearly demonstrate the robustness and generalization capability of our approach.

\subsubsection{Token-Level Disentangling Inspection}
\label{sec:token-level_disentangleInspection}

\begin{table}[t]
\small
\centering

\resizebox{0.45\textwidth}{!}{%
\begin{tabular}{@{}lcccccc@{}}
\toprule
 & capital & num. & punc. & Ar   & Ru   & Zh   \\ \hline
overall                & 5.44    & 3.18   & 0.25 & 34.2 & 37.2 & 22.1 \\
$\Delta_{acc} > 0$ & 5.68    & 3.34   & 0.37 & 33.9 & 40.8 & 18.5 \\
\bottomrule
\end{tabular}%
}
\caption{Results of Token-level Disentangle Inspection}
\label{tab:token-level_disentangleInspection}
\end{table}

In this subsection, we mainly discuss the effect of our proposed framework on token-level positional information disentanglement, and the advantages of our model compared to previous work.

To investigate the question, based on our analysis for the MNMT without PE in \S~\ref{sec:analysis}, we conduct token-level probing on both MNMT-RC and MNMT-TPDM models. For this probing task, we focus on tokens that preserve more positional information under our model. To be specific, we measure the proportion of different types of tokens in the entire vocabulary, and the proportion of different types of tokens with $\Delta_{acc } = ACC_{TPDM} - ACC_{RC} > 0$. The result is shown in Table.~\ref{tab:token-level_disentangleInspection}. 

According to the results, we find out that: (1) the proportion of punctuation, number, and tokens capitalized with the first letter increase significantly. This shows that our method successfully helps tokens to get access to useful positional information; (2) the proportion of tokens in all languages, except Russian, decreases. We hypothesize that the increase in the proportion of Russian tokens may be partly due to the presence of many capitalized tokens in Russian. In addition, as shown in Fig.~\ref{fig:bleu_acc}, our model can disentangle more positional information during training, which leads to a faster convergence during training and better results. We additionally provide some case studies in Appendix~\ref{apd:caseStudy} to give an intuitive impression of the observations. From the case studies, we find out that the MNMT-RC conspicuously suffers from the erroneous translation of number, and factual errors caused by the numerical mismatch. One possible reason is that it is due to the neighbor tokens' influence caused by positional information missing.

\section{Related Work}

Previous MNMT works have shown prominent capacity in zero-shot direction translation. \cite{johnson-etal-2017-googles} introduces an artificial language token as an auxiliary input to enable a single model to achieve multiple-pair machine translation and improve the zero-shot translation performance. \cite{lu-etal-2018-neural} further verifies the massive multilingual translation model's feasibility, especially in the zero-shot direction. However, there is still room for improvement in these methods. Previous research works~\cite{zhang-etal-2020-improving, al-shedivat-parikh-2019-consistency, gu-etal-2019-improved} have pointed out the destabilization and common off-target translation issue in zero-shot translation using the existing models. 

Based on these observations, meaningful research works have been published to tackle these issues from three major aspects. First, \cite{raganato-etal-2021-empirical, gu-etal-2019-improved, liu-etal-2021-improving-zero, blackwood-etal-2018-multilingual, sachan-neubig-2018-parameter} introduces innovative structures to learn language-agonistic representation and eliminate the source language-pair related bias. Second, well-designed data augmentation procedures are proposed by \cite{wang-etal-2021-rethinking-zero, gu-etal-2019-improved, zhang-etal-2020-improving}. Third, innovated regularization objectives are tried by existing research work \cite{arivazhagan2019missing, al-shedivat-parikh-2019-consistency}. Our work proposes a solution to improve the zero-shot MNMT performance by introducing a token-level position disentangle module. Recently, more work has tried to solve the problem from the latent variable perspective\cite{gu-etal-2019-improved, wang-etal-2021-rethinking-zero}. For those that are related to our work, \cite{liu-etal-2021-improving-zero} views positional information as a latent variable that zero-shot MNMT performance suffers from but leaves a blank of analysis about how positional information influences MNMT, especially in the zero-shot direction, systematically. Our work proposes further analysis and a more efficient mechanism accordingly.

\section{Conclusion}

The paper analyzes the specific mechanism regarding the influence of positional information on machine translation. We consider that the positional information is optionally needed for the MNMT system. Therefore, we propose TPDM to disentangle positional information at the token level and selectively remove the positional information for MNMT. Our experiments show that our proposed framework outperform MNMT-RC on the zero-shot directions, while reducing the performance loss on the supervised direction. Furthermore, compared to MNMT-RC, our proposed model can disentangle more positional information, leading to a faster convergence during training. Finally, in terms of the removing of token-level positional information, we find out that our proposed method can remove positional information well, while selectively preserve it for tokens indicating sentence-ordering (\textit{e.g.}, capital, punc.), or requiring strict order (\textit{e.g.}, number).

\section{Limitations}

The proposed MNMT-TPDM in this paper requires about 1\% additional parameters. In addition, multiple training objectives are applied during training. Both of these modification lead to the increase of algorithmic complexity and computational costs. We have revealed the number of the parameters and the time cost for training the model in Appendix~\ref{sec:efficiency}.

\bibliography{anthology,custom}
\bibliographystyle{acl_natbib}

\clearpage
\appendix

\begin{figure*}[t]
\centering
\includegraphics[width=0.8\textwidth]{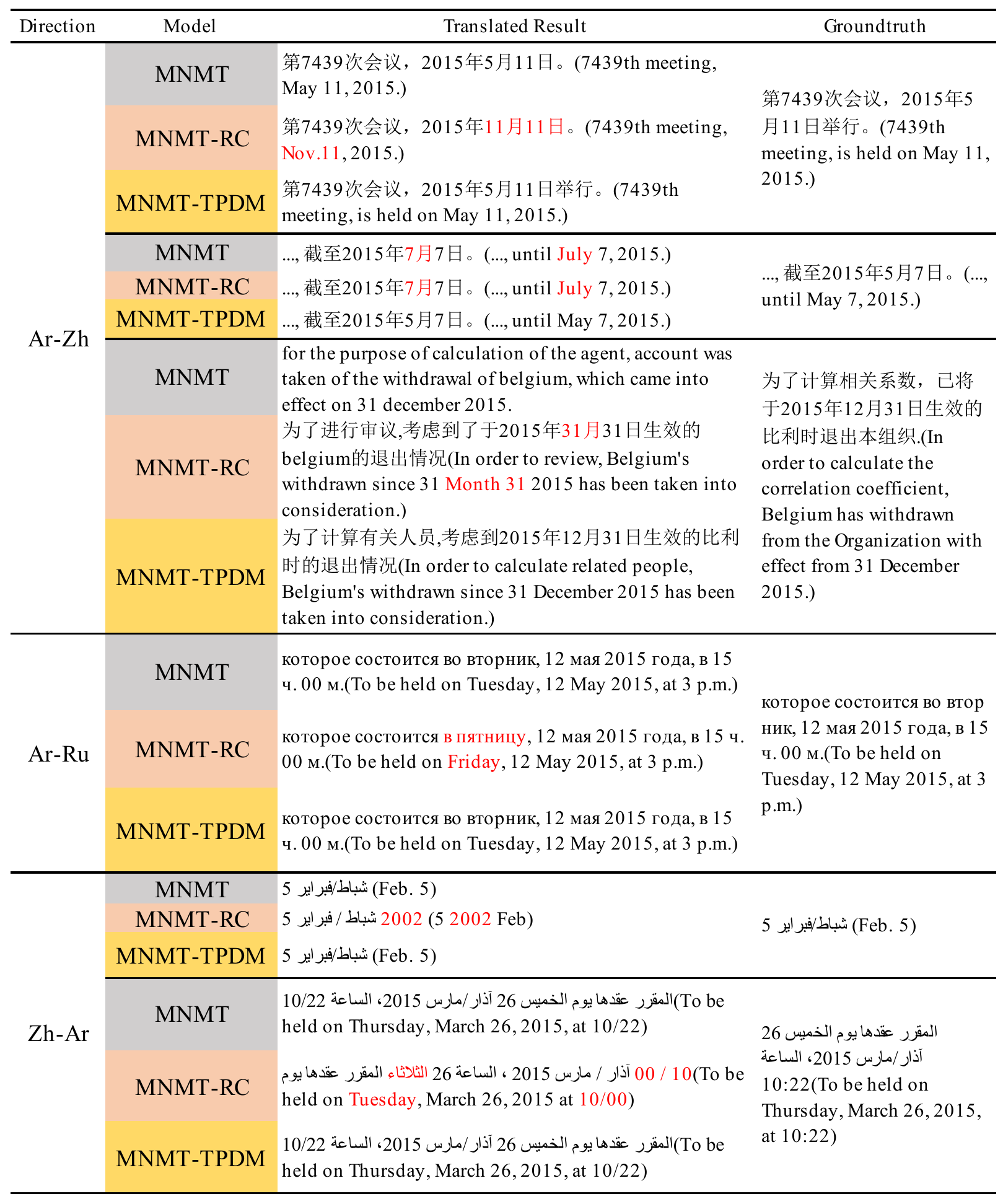}
% \vspace{-0.3in}
\caption{Case study on zero-shot directions.}
\label{fig:casestudy}
\end{figure*}

\section{Case Study}
\label{apd:caseStudy}

In this section, we offer several cases for further investigation. The cases are shown in Fig.~\ref{fig:casestudy}. From the cases, we find that MNMT-RC conspicuously suffers from number translation errors and factual errors caused by numerical mismatch. We believe that it is due to the neighbor tokens' influence caused by the positional information missing.

% \section{Full Experimental Results}

% \section{Adaptability of Position Disentangle Module}

\section{Dataset Statistics}

In this section, we show the dataset statistics of each dataset we used for training and evaluating the MNMT model. In our experiment, we sub-sample 2M data for MultiUN, and use all the data in Europarl. The statistics indicate the number of cleaned, segmented sentences that are used for training. Note that each number within the table contains data of all directions. In terms of Europarl, we follow \cite{wang-etal-2021-rethinking-zero} to utilize the \textit{devtest} datasets as the valid dataset to select the best model, and evaluate the model on the \textit{test} datasets.

\begin{table}[!h]
\centering
\begin{tabular}{lrrr}
\toprule
Dataset Statistics &  \multicolumn{1}{c}{train} & \multicolumn{1}{c}{valid} & \multicolumn{1}{c}{test} \\ \midrule
MultiUN            & 10703407                  & 6000                      & 48000                    \\
Europarl           & 7463628                   & 2000                      & 12000            \\ \bottomrule       
\end{tabular}
\caption{Dataset Statistics.}
\label{fig:statistics}
\end{table}

\begin{table*}[!th]
\centering
\resizebox{\textwidth}{!}{%
\begin{tabular}{@{}lccccccc|ccccccc}
\toprule
% \textbf{Task}                  & \multicolumn{7}{c}{\cellcolor{RBRed}MultiUN} &  \multicolumn{3}{c}{\cellcolor{RBBlue}Europarl} \\ \midrule & 
\textbf{Task}                  & \multicolumn{14}{c}{\cellcolor{RBRed}MultiUN} \\ \midrule & 
Ar$\rightarrow$En &	En$\rightarrow$Ar &	Ru$\rightarrow$En &	En$\rightarrow$Ru &	Zh$\rightarrow$En &	En$\rightarrow$Zh &	Supervised Avg. & Ar$\rightarrow$Zh  & Zh$\rightarrow$Ar & Ar$\rightarrow$Ru    & Ru$\rightarrow$Ar   & Zh$\rightarrow$Ru    & Ru$\rightarrow$Zh &	Zero-shot Avg. \\ \midrule
% Ar$\rightarrow$Zh  & Zh$\rightarrow$Ar & Ar$\rightarrow$Ru    & Ru$\rightarrow$Ar   & Zh$\rightarrow$Ru    & Ru$\rightarrow$Zh      & \textbf{Zero. Avg.} & De$\rightarrow$Fr  & Fr$\rightarrow$De  & \textbf{Zero. Avg.}     \\ \midrule
% \multicolumn{12}{l}{\textbf{Small Model}}   \\ \midrule
% PIV-M (small) & & & & & & & & 35.0 & 17.6 &  \\ 
% PIV-M (base)  & & & & & & & & 42.3 & 22.6  & 30.0   & 23.1  & 27.8   & 40.3 & 31.0 & 28.0  & 22.3 & 25.2 \\ \midrule
MNMT-small   &   48.1 & 32.5 & 45.6 & 36.3 & 43.2 & 51.2 & 42.8 & 7.2   & 4.7   & 4.1   & 5.4 & 3.9 & 7.1 & 5.4     \\
MNMT-RC-small & 47.7 & 32.1 & 45.1 & 36.0 & 42.8. & 51.1  & 42.5 & 10.3  & 6.9   & 4.2   & 6.5 & 2.8 & 10.5 & 6.9 \\
MNMT-TPDM-small & 47.7 & 31.9 & 45.3 & 35.7 & 42.7 & 50.6 & 42.3  & 26.2 & 13.2  & 18.7   & 14.3  & 16.6  & 26.0     & 19.2 \\ 
MNMT-GTPDM-small & 47.5 & 31.9 & 45.0 & 35.6 & 42.3 & 50.4 & 42.1 & 28.8 & 15.0 & 19.8 & 16.3 & 17.3 & 29.4 & 21.1 \\ \midrule 
% & \multicolumn{12}{l}{\textbf{Base Model}}   \\ \midrule  
% & \textbf{Supervised Avg.}  & ar-zh  & zh-ar & ar-ru    & ru-ar   & zh-ru    & ru-zh      & \textbf{Zero-shot Avg.}  & \textbf{Supervised Avg.} & de-fr  & fr-de  & \textbf{Zero-shot Avg.}      \\ \midrule
MNMT-base     &  53.9 & 36.7 & 50.1 & 42.2 & 49.9 & 57.1 & 48.3 & 29.3 & 15.0  & 20.9   & 16.2  & 18.6   & 29.4 & 21.6   \\
MNMT-RC-base   & 52.9 & 36.2 & 49.5 & 41.0 & 48.8 & 56.0 & 47.4     & 41.0 & 21.0  & 29.1  & 21.7   & 26.2   & 40.0 & 29.8 \\
MNMT-TPDM-base   & 53.2 & 36.4 & 49.7 & 41.7 & 49.3 & 56.7 & 47.8     & 44.0 & 22.4  & 30.7   & 23.1  & 27.7  & 42.7     & 31.8 \\
MNMT-GTPDM-base & 53.3 & 36.1 & 49.7 & 40.8 & 48.6 & 56.1 & 47.4 & 43.7 & 22.3 & 30.8 & 23.4 & 27.4 & 43.1 & 31.8 \\
\bottomrule
\end{tabular}%
}
\caption{Full experimental results of MultiUN.}
\label{tab:multiun}
\end{table*}

\begin{table*}[!th]
\centering
\resizebox{\textwidth}{!}{%
\begin{tabular}{@{}lccccc|ccc}
\toprule
% \textbf{Task}                  & \multicolumn{7}{c}{\cellcolor{RBRed}MultiUN} &  \multicolumn{3}{c}{\cellcolor{RBBlue}Europarl} \\ \midrule & 
\textbf{Task}                  & \multicolumn{8}{c}{\cellcolor{RBBlue}Europarl} \\ \midrule & 
De$\rightarrow$En &	En$\rightarrow$De &	Fr$\rightarrow$En &	En$\rightarrow$Fr &	Supervised Avg.  & De$\rightarrow$Fr    & Fr$\rightarrow$De &	Zero-shot Avg. \\ \midrule
% Ar$\rightarrow$Zh  & Zh$\rightarrow$Ar & Ar$\rightarrow$Ru    & Ru$\rightarrow$Ar   & Zh$\rightarrow$Ru    & Ru$\rightarrow$Zh      & \textbf{Zero. Avg.} & De$\rightarrow$Fr  & Fr$\rightarrow$De  & \textbf{ZerFr Avg.}     \\ \midrule
% \multicolumn{12}{l}{\textbf{Small Model}}   \\ \midrule
% PIV-M (small) & 23.7 & 19.2  & 20.7 & 33.3 & 24.9 & 24.7  & 20.0 & 22.4  \\ 
% PIV-M (base)  & 30.0   & 23.1  & 27.8   & 40.3 & 31.0 & 28.0  & 22.3 & 25.2 \\ \midrule
MNMT-small      & 30.3   & 24.6 & 33.8 & 33.7 & 30.6 & 15.3 & 9.4 & 12.4     \\
MNMT-RC-small   & 30.3   & 24.3 & 33.6 & 33.5 & 30.4 & 20.6 & 4.5 & 12.6 \\
MNMT-TPDM-small  & 30.4   & 24.2  & 33.7  & 33.5     & 30.5 & 21.8 & 17.9 & 19.9 \\ 
MNMT-GTPDM-small &  29.7 & 23.8 & 33.3 & 33.0 & 30.0 & 23.7 & 19.1 & 21.4 \\ \midrule 
% & \multicolumn{12}{l}{\textbf{Base Model}}   \\ \midrule  
% & \textbf{Supervised Avg.}  & ar-zh  & zh-ar & ar-ru    & ru-ar   & zh-ru    & ru-zh      & \textbf{Zero-shot Avg.}  & \textbf{Supervised Avg.} & de-fr  & fr-de  & \textbf{Zero-shot Avg.}      \\ \midrule
MNMT-base           & 33.7   & 27.1  & 36.7   & 36.3 & 33.5 & 27.4  & 21.2 & 24.3    \\
MNMT-RC-base        & 32.9  & 26.4   & 36.3   & 35.9 & 32.9 & 29.3 & 23.2 & 26.3 \\
MNMT-TPDM-base      & 33.0   & 26.6  & 36.1 & 36.0  & 33.0 & 29.2 & 24.0 & 26.6 \\
MNMT-GTPDM-base     & 32.9 & 26.0 & 35.9 & 35.9 & 32.7 & 29.9 & 24.3 & 27.1 \\ \bottomrule
\end{tabular}%
}
\caption{Full experimental results of Europarl.}
\label{tab:europarl}
\end{table*}

\section{Full Experimental Results}
In this section, we reveal the full experimental results of Table~\ref{tab:main_results} in Table~\ref{tab:multiun} and Table~\ref{tab:europarl}.

\section{Efficiency of TPDM}
\label{sec:efficiency}

\begin{table}[!tbp]
\begin{tabular}{lcc}
\toprule
                & \#Parameters & Time Cost \\ \hline
MNMT-small      & $30.781$M   & $1.0 \times$      \\
MNMT-TPDM-small & $31.046$M   & $1.46 \times$    \\
MNMT-base       & $94.642$M   & $1.0 \times$      \\
MNMT-TPDM-base  & $95.696$M   & $1.35 \times$    \\ \bottomrule
\end{tabular}
\caption{Statistics of the number of parameters and time cost.}
\label{tab:param}
\end{table}

In this section, we reveal the difference in parameter and training cost in terms of our proposed MNMT-TPDM. The statistics are shown in Table~\ref{tab:param}. From the table, the parameters of our proposed MNMT-TPDM have no more than $1\%$ increase on both base and small version of models. In terms of time cost, the training time of our proposed model is increased by around $40\%$. This is mainly due to the adversarial training procedure.

% \appendix

% \section{Example Appendix}
% \label{sec:appendix}

% This is an appendix.

\end{document}